\documentclass[conference]{IEEEtran}
\IEEEoverridecommandlockouts
\usepackage{cite}
\usepackage{amsmath,amssymb,amsfonts}
\usepackage{algorithmic}
\usepackage{graphicx}
\usepackage{textcomp}
\usepackage{xcolor}
\usepackage{subcaption}
\def\BibTeX{{\rm B\kern-.05em{\sc i\kern-.025em b}\kern-.08em
    T\kern-.1667em\lower.7ex\hbox{E}\kern-.125emX}}
\begin{document}

\title{Evaluating Vision-Language Models for Zero-Shot Detection, Classification, and Association of  Motorcycles, Passengers, and Helmets}

\author{\IEEEauthorblockN{Lucas Choi}
\IEEEauthorblockA{\textit{Archbishop Mitty} \\
lucasleechoi@gmail.com}
\and
\IEEEauthorblockN{Ross Greer}
\IEEEauthorblockA{\textit{University of California San Diego} \\
regreer@ucsd.edu}
}

\maketitle
\begin{abstract}
Motorcycle accidents pose significant risks, particularly when riders and passengers do not wear helmets. This study evaluates the efficacy of an advanced vision-language foundation model, OWLv2, in detecting and classifying various helmet-wearing statuses of motorcycle occupants using video data. We extend the dataset provided by the CVPR AI City Challenge and employ a cascaded model approach for detection and classification tasks, integrating OWLv2 and CNN models. The results highlight the potential of zero-shot learning to address challenges arising from incomplete and biased training datasets, demonstrating the usage of such models in detecting motorcycles, helmet usage, and occupant positions under varied conditions. We have achieved an average precision of 0.5324 for helmet detection and provided precision-recall curves detailing the detection and classification performance. Despite limitations such as low-resolution data and poor visibility, our research shows promising advancements in automated vehicle safety and traffic safety enforcement systems.
\end{abstract}

\begin{IEEEkeywords}
Vehicle safety, Zero-shot learning, Vision-language models, Helmet detection, Traffic enforcement systems
\end{IEEEkeywords}

%%%%%%%%% BODY TEXT
\section{Introduction}

Motorcycle accidents are frequent causes of injury and death worldwide, especially for occupants not wearing helmets \cite{abdi2022motorcycle, souza2020mortality, abedi2017epidemiological, cavalcanti2013motorcycle}. 
Specifically, in India, in 2022, two-wheeler deaths accounted for 44\% of total road fatalities with 74,897 deaths, the highest out of all modes of transport\footnote{https://opencity.in/analysing-the-morth-road-accidents-report-for-2022/}. Helmets are 35\% effective in reducing the risk of Abbreviated Injury Scale 3+ head injuries \cite{jayaramanhelmet}. Additionally, 4 people die every hour in India because they do not wear a helmet\footnote{https://www.indiatoday.in/diu/story/two-wheeler-death-road-accidents-helmets-states-india-1602794-2019-09-24}, causing 44,666 deaths in 2019 \cite{unhelmeted}.

Section 129 of the Motor Vehicles Act in India states that ``Every person ... on a motorcycle of any class or description shall, while in a public place, wear protective headgear conforming to such standards as may be prescribed by the Central Government." \cite{Ministry-of-Road-Transport-&-Highways} 
Despite regulations mandating helmet use, compliance is inconsistent, leading to preventable injuries.

Iterations of the CVPR AI City Challenge \cite{Shuo24AIC24} have prompted researchers to address this challenge, stating 
``Motorcycles are one of the most popular modes of transportation, particularly in developing countries such as India. Due to lesser protection compared to cars and other standard vehicles, motorcycle riders are exposed to a greater risk of crashes. Therefore, wearing helmets for motorcycle riders is mandatory as per traffic rules, and automatic detection of motorcyclists without helmets is one of the critical tasks in enforcing strict regulatory traffic safety measures." We suggest that, besides the enforcement of traffic safety measures, there is also an even greater benefit in the ability of IoT-style communication between infrastructure or egocentric perception devices. Such systems could detect the presence of motorcyclists and passengers (with or without helmets) and alert the surrounding vehicles whose drivers (autonomous or human) may be otherwise unaware of the vulnerable road users in their proximity \cite{greer2023pedestrian}. 

Accordingly, to perceive holistic information about motorcycles and their occupants in a scene, the goal task we evaluate in this paper is the detection and classification of the following objects in every frame of a large video dataset:
\begin{enumerate}
    \item Motorcycle,
    \item Drivers wearing helmets,
    \item Drivers not wearing helmets,
    \item Passengers wearing helmets,
    \item Passengers not wearing helmets,
    \item 2nd Passengers wearing helmets,
    \item 2nd Passengers not wearing helmets,
    \item Children sitting in front of the driver wearing helmets,
    \item Children sitting in front of the driver not wearing helmets.
\end{enumerate}

\begin{figure*}
    \centering
\includegraphics[width=.20\textwidth, height=3.4cm]{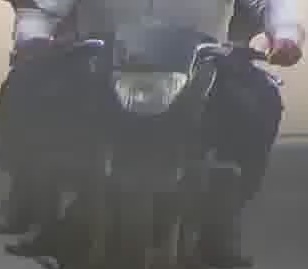}
\includegraphics[width=.11\textwidth, height=3.4cm]{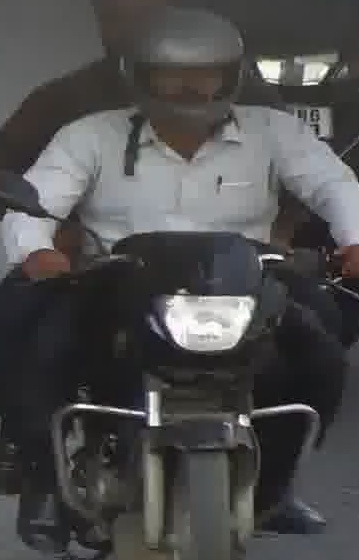}
\includegraphics[width=.11\textwidth, height=3.4cm]{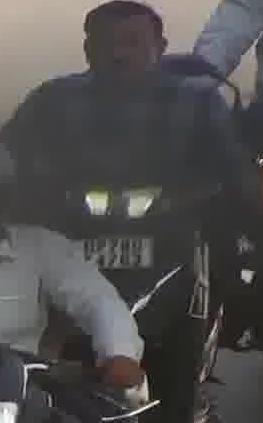}
\includegraphics[width=.105\textwidth, height=3.4cm]{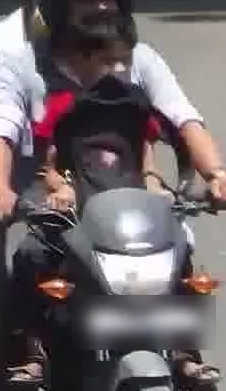}
\includegraphics[width=.14\textwidth, height=3.4cm]{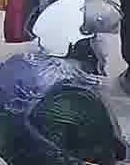}
\includegraphics[width=.125\textwidth, height=3.4cm]{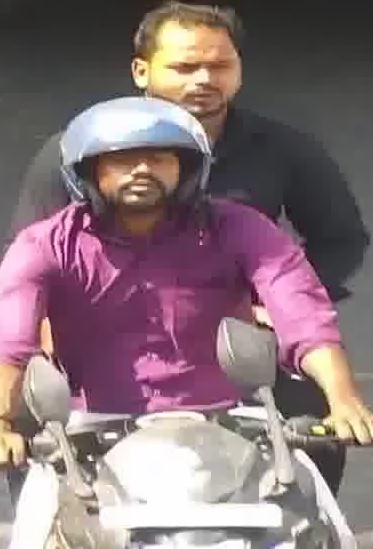}
\includegraphics[width=.16\textwidth, height=3.4cm]{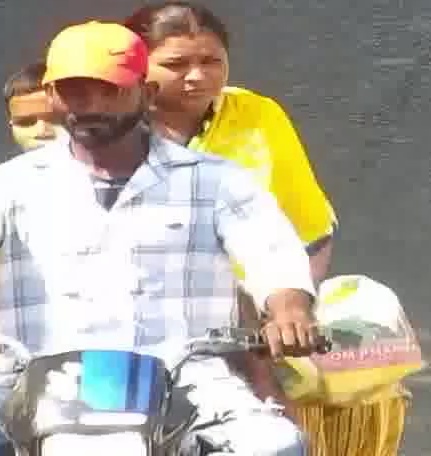}
    
    \caption{Example instances of classes to detect, cropped from the AI City Challenge dataset. From left to right: Motorcycle, Driver with Helmet, Driver with No Helmet, Child Passenger with No Helmet, Passenger 1 with Helmet, Passenger 1 with No Helmet, Passenger 2 with No Helmet.}
    \label{fig:exampleinstances}
\end{figure*}

In this research dataset, these scenes are captured by infrastructure-mounted cameras, though the same models can also be applied to egocentric views. This is especially the case given the zero-shot learning approaches we take, which do not require specific-view training data to be applied. We show sample data of these classes in Figure \ref{fig:exampleinstances}. 

With many data-driven applications, a common challenge is the ability of a training set to adequately represent the diversity of instances that appear in the real world \cite{ghita2024activeanno3d, greer2024and}. For this reason, data-driven methods excel when given the most data possible, as this increases the likelihood of learning similar patterns to a real-world instance. To this end, we create a method that extends beyond the dataset presented by Shuo et al. \cite{Shuo24AIC24} by employing a pre-trained vision-language foundation model for this detection task, specifically, the OWLv2 \cite{minderer2023scaling}. Further, in our research, we present a strategy for cascading models to modularly isolate and improve task performance for these important safety systems.

This foundation model strategy is important especially in consideration of challenges presented by dataset shortcomings. The given dataset has no instances of a child passenger with a helmet or a second passenger with a helmet. This is a huge hindrance in accurately detecting the seat position and helmet status in these specific classes using traditional machine-learning approaches due to the lack of data for training. Therefore, the use of zero-shot learning may provide a means to identify these instances in `real world' test data even without specific training. 

The question we explore in this research is to what degree such foundation model approaches, namely OWLv2, are ready for use with real-world data in this motorcycle safety road scene perception domain and where their strengths and weaknesses may lie. 

\section{Related Research}

Conventional machine learning object detection algorithms rely on manual annotations and specialized algorithms, which can be time-consuming and resource-intensive to label, especially as the models are limited to learning from provided datasets. Moreover, these methods often lack the flexibility to adapt to new environments or variations in helmet designs \cite{Object-Detection-Review}. 

Foundation models, with billions of parameters trained on enormous collections of information, have recently led to effective zero-shot techniques for a variety of tasks \cite{greer2024towards}, where a learned model can provide strong performance on datasets unseen during training \cite{radford2021learning}. One such foundation model is OWL-ViT \cite{minderer2022simple}; OWL stands for ``open-world localization", referring to this model's ability to function in an ``open" world (i.e., non-rigidly specified set of expected classes). The ViT portion of OWL-ViT refers to the Vision Transformer, an architecture that applies the attention mechanism to images instead of the prior standard of convolution. The OWL family of models uses contrastive learning between batches of image patch encodings and text embeddings, with image patch encodings producing proposed classes and proposed bounding boxes, and treating detection as a bipartite matching problem between these decoded image classes and bounding boxes, as introduced in the Detection Transformer (DETR) technique \cite{carion2020end, greer2022salience, greer2023salient}. Together, these methods were shown to be effective in zero-shot object detection (identifying a bounding box around desired classes of interest within an image). This method was refined and scaled up using self-training as OWLv2 \cite{minderer2023scaling}, whereby pseudo-box annotations are provided from an existing detector, and it is this further-trained model that we use in the method shared in this research.

For the same application of detecting and classifying the given objects detailed in Section I, many different approaches have been tried in the previous AI City Challenges; in the 2023 AI City Challenge \cite{Naphade23AIC23}, Tran et al.
\cite{Tran_2023_CVPR} used YOLOv8 for a score of 0.7754 for the mean average precision (mAP). Cui et al.
\cite{Cui_2023_CVPR} used DETA \cite{ouyang2022nms} ensemble and Detectron2 for a mAP of 0.8340.
In the 2024 AI City Challenge \cite{Shuo24AIC24}, mainly transformer models combined with ensemble techniques were used.
Vo et al. \cite{Vo_2024_CVPR} used Co-DETR \cite{zong2023detrs} with a Minority Optimizer for class imbalance and a Virtual Expander for a mAP of 0.4860.
Chen et al. \cite{Chen_2024_CVPR} used a DETA and DETR fusion model for a score of 0.4824 mAP.

\section{Algorithms for Image Processing with Vision-Language Detection}

To address the challenges of accurately detecting and classifying motorcycles, their passengers, and helmet usage, we developed a cascading detection algorithm for OWLv2. Furthermore, due to OWLv2's shortcomings, we employed an AlexNet for the seat classification task. This section outlines our cascading detection algorithm using OWLv2 and discusses our approach for the seat classification task.

We first note that there are abstract classes that relate the target classes to one another; for example, ``motorcycle" and ``person" are the abstract classes represented in the data scheme, where ``person" can be further classified based on the attributes of helmet-wearing and seating position. Due to this, our first goal is to detect these high-level classes. Further, we know that there is no driver or passenger without a motorcycle, so we only detect ``person" in association with a particular motorcycle instance.

\begin{figure*}
    \centering
    \includegraphics[width=0.9\textwidth]{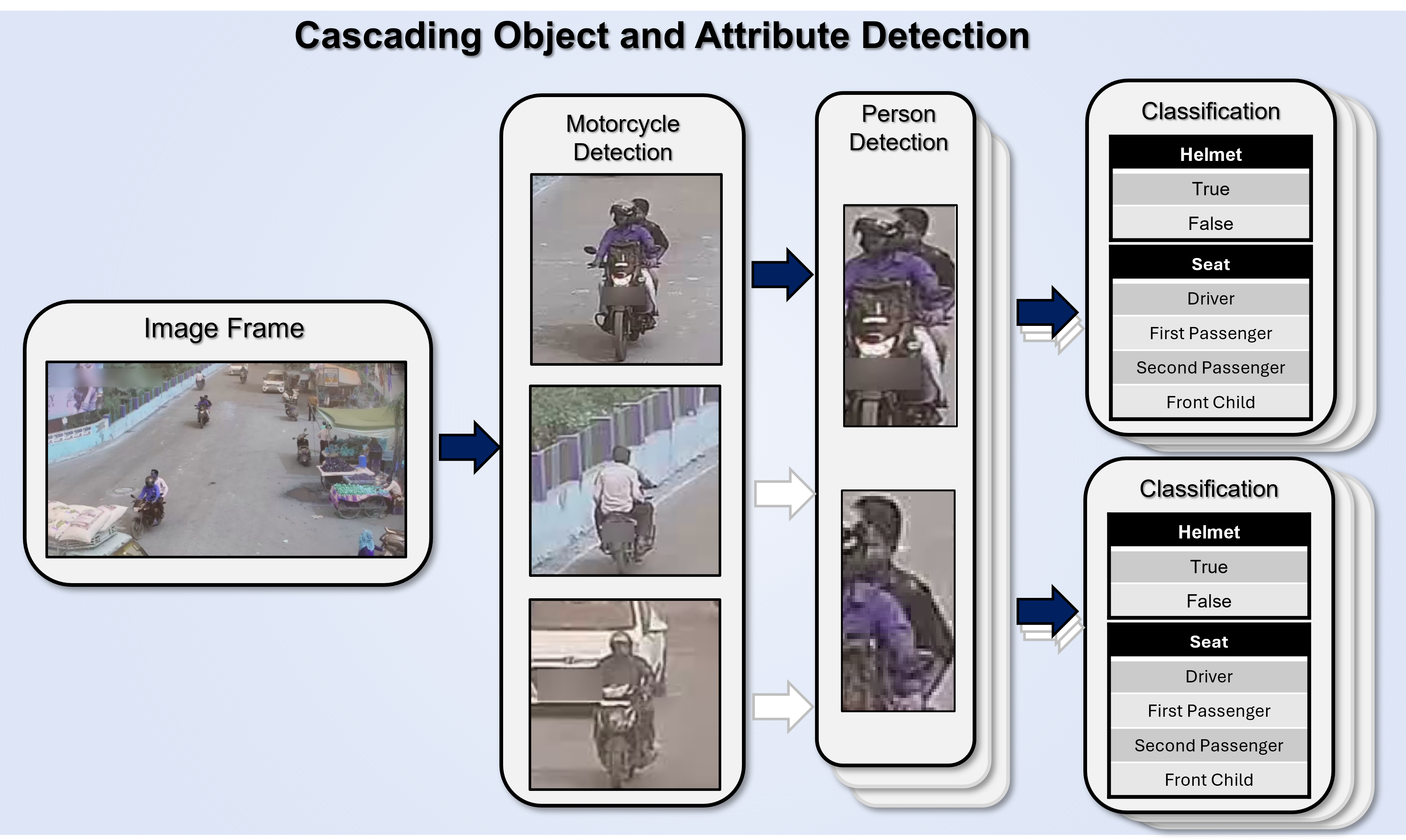}
    \caption{Our algorithm for detecting the relevant objects for helmet safety, as well as the appropriate attributes, acts in a cascaded style. First, from the original image, we detect all motorcycles. Then, within each motorcycle, we detect all human occupants (drivers and passengers). Then, for each detected human, we perform helmet detection and seat position classification. All detections, including helmet detection for the purpose of classification, are done using OWLViT2, while seat position classification is done using AlexNet.}
    \label{fig:enter-label}
\end{figure*}

Our detection algorithm, illustrated in Figure \ref{fig:enter-label}, begins with a detection stage. We provide a scene image (resized to 960 by 960 pixels and values normalized in [0, 1], relative to the size of each individual image in the batch) as input to OWLv2 along with the text ``motorcycle". The CLIPTokenizer, from \cite{radford2021learning}, encodes the text to be wrapped by the processor with the normalized image. 

%Using OWLv2, first described in \cite{minderer2023scaling}, we tested the model on the first frame of the first video. We detected the bounding boxes of the motorcycles, testing the model's functions. OWLv2 scales down the image to a resolution of 960x960 by default. We printed the re-scaled bounding boxes after calculating the scale to the original size. 

To detect the person instances on the motorcycles, we expand the re-scaled bounding box by 50 pixels on the left, right, and top sides to encapsulate any person instances surrounding the motorcycle. Using the expanded box, we crop the original image and run the OWLV2 model over this cropped image with the prompt ``person" to detect person instances. 
%Given the new bounding boxes, we re-scaled them to the cropped image and transformed them with the coordinates of the cropped image to make the bounding box coordinates in relation to the original image.

The algorithm's subsequent step is to perform the next level of detection, focusing on helmets, by cropping each person instance and running the OWLv2 over the cropped image with a text input of ``helmet". Because our task is to classify each person based on whether they are wearing a helmet and not necessarily to detect the helmet itself, we store the boolean result of this detection as an attribute of the person (rather than noting the bounding box).

We note that there is a general difficulty of OWLv2 in differentiating a person's semantic position on the motorcycle (such as driver, passenger, second passenger, etc.), as noted in Section IV. For this particular portion of the task, we take a supervised learning approach. We seek to provide each person detected on the motorcycle with an attribute of location between the positions enumerated in the introduction. 

Therefore, we use a neural network (a variant of AlexNet \cite{krizhevsky2012imagenet}, with a final layer output of four) to classify the seating position on a motorcycle of the person instances detected with OWLv2. 

Due to the use of the AlexNet in the seat position classification task, we recognize that the whole process is not completely zero-shot. It is rather a hybrid of zero-shot learning and supervised learning, with zero-shot for the association and detection of motorcycles, their passengers, and their helmet status, and supervised learning for the seat classification of the passengers. In this way, the methods in this paper actually address four tasks (motorcycle detection, person detection, helmet detection, and seat classification); three of these are solved in a zero-shot manner, and we include a learned approach to seat classification as this is a relevant safety task that should also be considered in conjunction.

In total, this algorithmic sequence of tasks can provide detections of motorcycles, associated people, their positions, and their helmet status for each image in a video.

% To process the training data, we iterate the labels and ground truth text files with the training videos, cropping each frame for each object within the ground truth file and sorting them into folders named by the labels. While training the neural network, we excluded the motorcycle detections. With this data, we trained the neural networks to detect what category each cropped image of people we input falls into.

\section{Experimental Method and Evaluation}

Using the cascaded object and attribution detection algorithm detailed in the previous section, we performed detection on the dataset of 100 videos provided by \cite{Shuo24AIC24}, with further implementation details described in this section.

%We first processed each frame individually with OWLv2, detecting the motorcycles using inputs as described in Section III 
%We compounded our detections in the same frame using our cascaded model format. As described in Section III, we subsequently detect the person inside the cropped motorcycle bounding box by setting the text as "person". 
%Then, we repeat this process with the helmet with an input of "helmet", detecting it within the person's bounding box. 

We first conducted the motorcycle and person detection of our cascaded detection process as described in Section III. 

The threshold for OWLv2 is a confidence threshold, meaning it is the minimum confidence score that a predicted bounding box must have to be considered a valid detection. The OWLv2 will discard any detection with a confidence score below the given threshold. The confidence score is calculated as logits on a per-detection basis.

We performed the cascaded detection process with thresholds of 0.1 to 0.7 on the OWLv2 to examine the sensitivity of precision and recall to thresholds. 0.7 was chosen as the last threshold, as OWLv2 made no detections with a threshold higher than 0.7. Using the output of our detections, we calculated the precision and recall at each confidence threshold. 

To evaluate the ability of OWLv2 to classify a passenger's helmet status, regardless of error in upstream person detection, we detected helmets within the ground truth bounding boxes of passengers to classify the passenger's helmet status. 
As in the previous detections, we experimented with a threshold of 0.05 to 0.7. 
%As the helmet dataset is relatively smaller; we had the resources to lower the threshold limit.

When performing the seat classification based on the person detection, we attempted to determine a passenger's seat with OWLV2, first using the text prompts provided by the labels in the dataset, such as ``passenger 1" and ``child passenger." However, with these prompts, OWLv2 tended to miss some passengers and mislabel the people. Assuming this was due to the inputs, we attempted more specific prompts such as ``child in front of driver" or ``passenger behind driver". Nevertheless, this also yielded similar results. We hypothesize that the prompt inputs were not the determining factor of OWLv2's failure to detect and differentiate the different people on a motorcycle, showing possible shortcomings of model training for this particular type of task. Furthermore, the task of classifying people based on their relative location to other people and the motorcycle may be too specific for the model.

After observing OWLv2's shortcomings with our intersection data, we used a modified AlexNet for the seat classification subtask \cite{krizhevsky2012imagenet}. We modified the last layer of the AlexNet from 10 outputs to 4 to suit our task. 

We used an approximate inverse class frequency to overcome the severe class imbalance in the dataset as shown in Table \ref{tab:GroundTruth}. At first, we tested the weighting of 1.147, 7.908, 785.229, and 2093.944, calculated by inverse class weighting. However, this was insufficient, as the model did not appear to learn the child class and appeared to over-favor the driver class. Therefore, we incrementally increased the weighting of the classes of passenger1, passenger2, and child passenger relative to the driver, updating the previously mentioned weights to 1, 10, 800, and 3000, respectively. 

We split the data 70/15/15 for the training, testing, and validation. We used a cross-entropy loss. Finally, we trained the model using a learning rate of 0.0001 for 100 epochs. Then, we used the model with the lowest loss on the validation set to make inferences on the test set. % and achieved the best loss on the twelfth epoch.

\begin{table}[]
    \centering
    \caption{Ground Truth Data}
    \begin{tabular}{c|c}
        
        Class & Frequency \\
        \hline \hline
        Driver & 32,889 \\
        \hline
        Passenger 1 & 4,796 \\
        \hline 
        Passenger 2 & 78 \\
        \hline 
        Child Passenger & 48\\
        \hline  \hline
        Total & 37,811\\
        
    \end{tabular}
    \label{tab:GroundTruth}
\end{table}

\subsection{Data}

\begin{figure}

  \centering
\includegraphics[width=.4\textwidth]{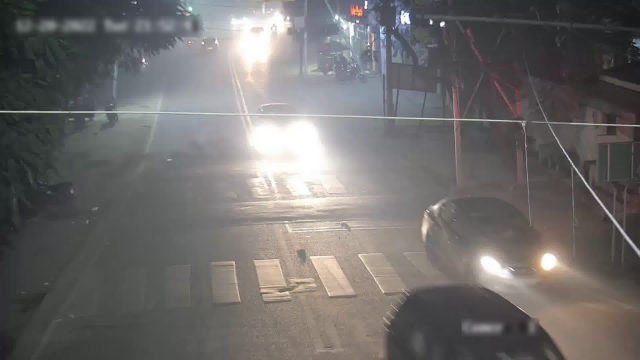}
\includegraphics[width=.4\textwidth]{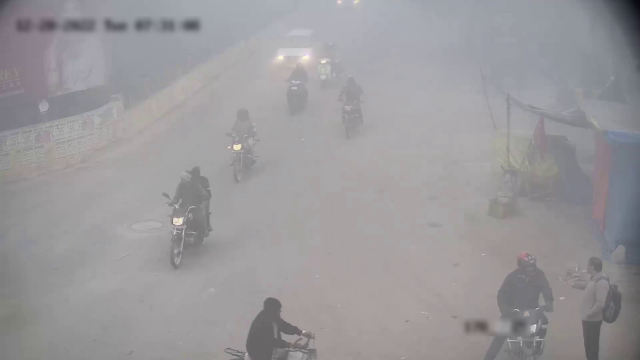}
\includegraphics[width=.4\textwidth]{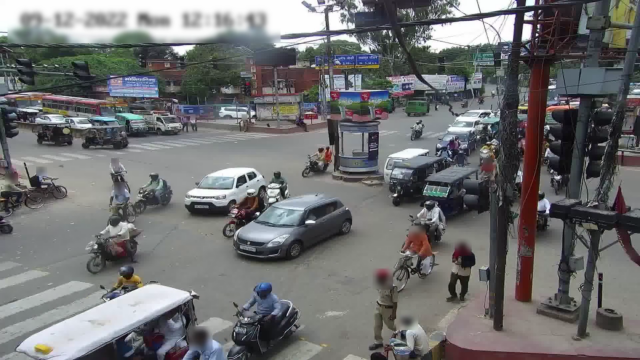}
\caption{Sample images of the dataset of different angles with different environments. From top to bottom: night, foggy, crowded}
\label{fig:images}
\end{figure}

The dataset provided by \cite{Shuo24AIC24} contains 100 videos taken by infrastructure-mounted cameras in India. They are annotated with bounding boxes of motorcycles and up to four passengers who may or may not be wearing helmets. Each video is 20 seconds long, sampled at 10 Hz, and has a resolution of 1920×1080. Example images from the dataset are shown in Figure \ref{fig:images}. The ground truth data is comprised of class frequencies, as shown in Table \ref{tab:GroundTruth}, and has 26349 helmet-wearers and 11462 unhelmeted people, meaning 69.7\% are helmeted.

% \begin{figure}
% \begin{subfigure}{.5\textwidth}
%   \centering
%   \includegraphics[width=7.1111cm, height = 4cm]{night1.png}
%   \caption{Night}
%   \label{fig:sfig1}
% \end{subfigure}%
% \begin{subfigure}{.5\textwidth}
%   \centering
%   \includegraphics[width=7.1111cm, height = 4cm]{fog1.png}
%   \caption{Fog}
%   \label{fig:sfig2}
% \end{subfigure}
% \begin{subfigure}{.5\textwidth}
%   \centering
%   \includegraphics[width=7.1111cm, height = 4cm]{overlap.png}
%   \caption{Crowded}
%   \label{fig:sfig2}
% \end{subfigure}
% \caption{Sample images of the dataset of different angles with different environments}
% \label{fig:fig}
% \end{figure}

\subsection{Results}
\begin{figure}
    \centering
    \includegraphics[width=.4\textwidth]{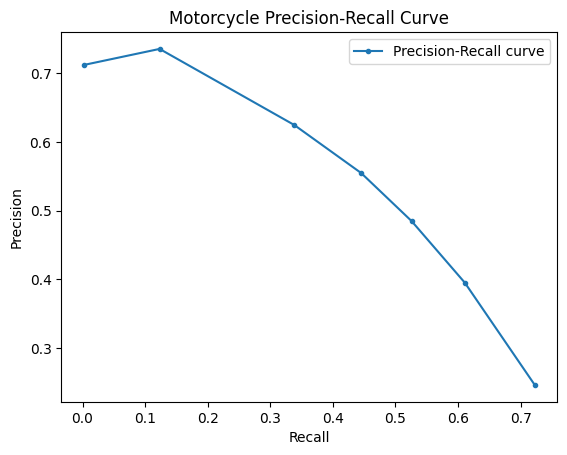}
    \caption{Precision-Recall Curve of Motorcycle Detection. Initially, a slight increase in precision indicates improved confidence in early predictions. However, precision declines steeply as recall rises, highlighting the model’s challenge in maintaining accuracy while capturing more true positives. }
    \label{fig:motorcycle}
\end{figure}

\begin{figure}
    \centering
    \includegraphics[width=.4\textwidth]{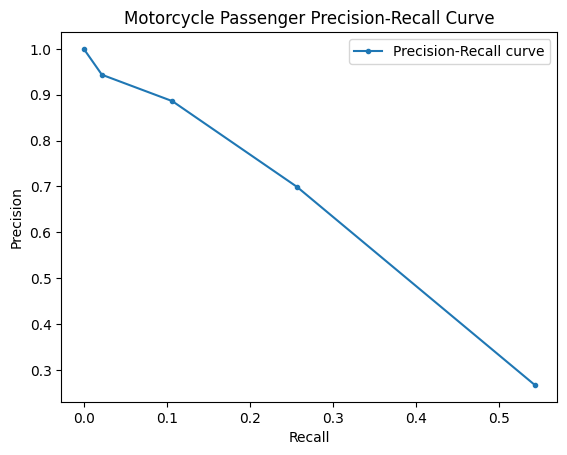}
    \caption{Precision-Recall Curve of Passenger Detection.     The curve demonstrates a high precision at low recall values. Despite the trade-off of precision and recall, the shape suggests a robust model performance in balancing the two. }
    \label{fig:personAcc}
\end{figure}

Our OWLv2 detected motorcycles with accuracies shown in Table \ref{tab:AccuraciesMotorcycle} and detected persons with accuracies as shown in Table \ref{tab:AccuraciesPerson}. For motorcycle detection, the average precision is 0.4122, calculated by the area under the curve of Figure \ref{fig:motorcycle}, and for person detection, the average precision is 0.3561, obtained from Figure \ref{fig:personAcc}.

\begin{figure}
    \centering
    \includegraphics[width=.4\textwidth]{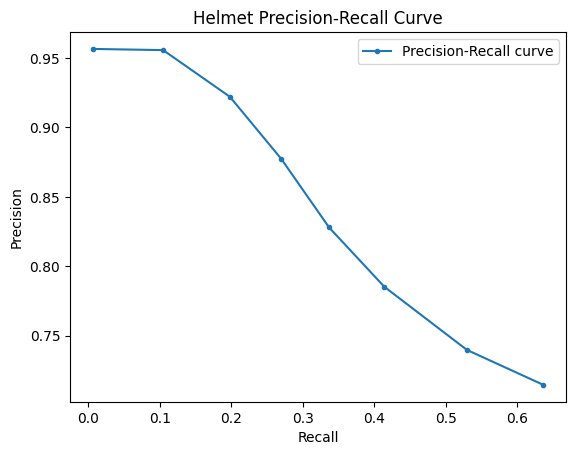}
    \caption{Precision-Recall Curve of Helmet Classification \\
    The curve has a high initial precision, progressively decreasing as recall increases. Efforts to capture more true positives resulted in a higher incidence of false positives.}
    \label{fig:helmAcc}
\end{figure}

Our helmet-status classification was done through helmet detection with OWLV2 on the provided ground truth bounding boxes of passengers, with a representative classification based on the helmet's presence or absence. This resulted in the precisions and recalls in Table \ref{tab:AccuraciesHelmet}, tested over multiple thresholds, resulting in an average precision of 0.5324, as further illustrated in Figure \ref{fig:helmAcc}. A naive classifier, which always predicts the rider to be wearing a helmet, would have a precision of 69.7\% and a trivial recall of 100\% based on the ground truth data described in Section IV A; at all thresholds, our precision is higher than the naive classifier, showing a reduction in false positives and negatives.

The IoU in Tables \ref{tab:AccuraciesMotorcycle}, \ref{tab:AccuraciesPerson}, and \ref{tab:AccuraciesHelmet} stands for intersection over union, which is the metric for evaluating the accuracy of a predicted bounding box. The IoU is calculated as follows: \begin{math}
IoU=\frac{A\cap B}{A\cup B}
\end{math} or \begin{math}
IoU=\frac{Area\:of\:Overlap}{Area\:of\:Union}
\end{math}, where \begin{math}A\end{math} stands for the predicted bounding box and \begin{math}B\end{math} stands for the ground truth bounding box.
In our evaluation, for a given detected bounding box, if the IoU with the ground truth is greater than or equal to 0.5, then the detection is considered a ``true positive".

% a loss of 872.071 and
Finally, our neural network's seat classification achieved an accuracy of 95.17\% on the validation set, with the classification results on the test set displayed in Figure 3. We note that the severe class imbalance does leave the child passenger class unsuccessfully classified, though this does not have much impact on the accuracy metric. This reveals an insufficiency in the model learning and cautions us of evaluating performance for such an imbalanced dataset without examining class performance in the confusion matrix. 

\begin{table}[]
    \centering
    \caption{Precision and Recall scores of OWLv2 Motorcycle detections. No detections were made above the threshold of 0.7.}
    \begin{tabular}{c|c|c}
        Threshold & Precision (IoU 0.5) & Recall (IoU 0.5) \\
        \hline
        0.7 & 0.7124 & 0.002095\\
        0.6 & 0.7357 & 0.1232\\
        0.5 & 0.6249 & 0.3384\\
        0.4 & 0.5548 & 0.4453\\
        0.3 & 0.4849 & 0.5258\\
        0.2 & 0.3951 & 0.6108\\
        0.1 & 0.2460 & 0.7226\\
    \end{tabular}
    \label{tab:AccuraciesMotorcycle}
\end{table}

\begin{table}[]
    \centering
    \caption{Precision and Recall scores of OWLv2 Passenger Detection. No detections were made above the threshold of 0.5.}
    \begin{tabular}{c|c|c}
        Threshold & Precision (IoU 0.5) & Recall (IoU 0.5) \\
        \hline
        0.5 & 1.0 & 6.6136e-5 \\
        0.4 & 0.9437 & 0.02183 \\
        0.3 & 0.8861 & 0.1066\\
        0.2 & 0.6992 & 0.2568 \\
        0.1 & 0.2672 & 0.5432\\
    \end{tabular}
    \label{tab:AccuraciesPerson}
\end{table}

\begin{table}[]
    \centering
    \caption{Precision and Recall scores of OWLv2 Helmet Classification. No detections were made above the threshold of 0.7.}
    \begin{tabular}{c|c|c}
        Threshold & Precision (IoU 0.5) & Recall (IoU 0.5) \\
        \hline
        0.7 & 0.9565 & 0.005845 \\
        0.6 & 0.9557 & 0.1046 \\
        0.5 & 0.9221 & 0.1980 \\
        0.4 & 0.8775 & 0.2698\\
        0.3 & 0.8280 & 0.3370 \\
        0.2 & 0.7852 & 0.4143\\
        0.1 & 0.7398 & 0.5298 \\
        0.05 & 0.7146 & 0.6370\\
    \end{tabular}
    \label{tab:AccuraciesHelmet}
\end{table}

\begin{figure}
    \centering
    \includegraphics[width=.4\textwidth]{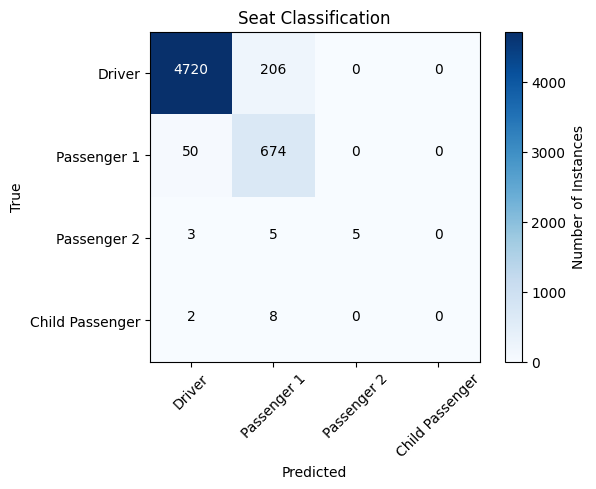}
    \caption{Confusion Matrix of Seating Position Classification}
    \label{fig:seats}
\end{figure}

\subsection{Sensitivity of Helmet Detection to OWLViT Detection Threshold}
Due to the nature of the dataset, it is important to find an optimal threshold in our detections. As many videos within the data are often unclear, too high of a threshold may omit the detections within the unclear regions of the data. On the contrary, too low of a threshold may yield unrelated detections, such as detecting a bike as a motorcycle. Therefore, an optimal threshold between these two extremes is necessary to achieve the highest accuracy. We show the results of exploring multiple thresholds in our research and note that continual tuning will be important when applying these methods to additional datasets or tasks.

\section{Concluding Remarks and Future Research}

Zero-shot learning demonstrates the potential of this application as it can overcome some limitations of incomplete and biased training datasets. As noted, the provided dataset lacks instances of child passengers with helmets and second passengers with helmets, making training traditional supervised-learning models difficult. Zero-shot learning leverages pre-training on diverse data, classifying unseen instances more accurately. With further fine-tuning and training, zero-shot learning has a strong potential for accurately handling real-world data.

Several sources of error are demonstrated in the helmet classification. The ground truth bounding boxes do not always encompass the whole person. Many boxes were taken from the lower half of their body as they entered the frame of the video. Additionally, overlapping bounding boxes with passengers and drivers, where drivers have helmets on, but the passengers do not, often confuses the OWLV2, claiming that it had detected the helmet in both cases. This also impacts person detection, as the OWLv2 cannot detect the passenger due to the driver obstructing most of the passenger's body.

Furthermore, AlexNet and the OWLv2 foundation model were challenged when faced with `real-world' noise-filled scenarios. Many of the videos provided in the dataset had very low resolutions, with blurred-out time stamps at the top left and bottom right obstructing the view of motorcycles. Data collected during the night further reduced visibility, as the headlights of motorcycles and cars create a blurry effect throughout the video. The regular poor conditions of fog or heavy air pollution compounded these factors, as shown in Figure \ref{fig:images}. All of these various aspects made image detection and classification challenging and sometimes near impossible. 

Future investigations are necessary to apply zero-shot learning in the real world. In this application, accurately detecting helmets will help to raise awareness as the detections will provide a more accurate measure of the frequency at which people do not wear helmets, as well as assist in enforcing the wearing of helmets. The ability to respond to unanticipated data is crucial for safety systems, as real-world scenarios often surpass the scope of any pre-existing dataset. Ongoing development and refinement of the model will be imperative to fully harness their potential in practical safety systems. Our future research will focus on enhancing the accuracy, robustness, and consistency of zero-shot learning models in our detections. 

To handle noisy data, pre-training the OWLv2 on further diverse datasets will allow it to better handle uncertain detections. Furthermore, preprocessing the data will mitigate some of these issues. Moreover, a possible improvement is the further integration of AlexNet and OWLv2 for seat classification. A hybrid approach using these two models will involve ensemble methods to balance their strengths for a more accurate result \cite{greer2023ensemble}.

Finally, we will address task-specific shortcomings. For example, at times, the OWLv2 model fails to get the bounding box over the whole person, specifically the head, which is especially crucial for this task. A primary focus will be improving the model's ability to localize and classify these critical areas accurately.

Despite current limitations and imbalances in data, this research shows the potential of foundation models and language-based prompting toward the zero-shot handling of important safety challenges. We address all components of the AI City Challenge Helmet Detection and Occupancy tasks, showing possibilities for the OWL model to address the sub-tasks of detection and association of vehicles, their occupants, and safety state information. This application has the potential to extend upon I2V communication. The detections from the infrastructure point of view can be sent to the vehicle's egocentric perception in order to alert drivers of the presence of motorcycles for safer intersection driving.

%As we continue to fine-tune the models to improve accuracies, we will also perform a more robust test over different thresholds to construct a precision-recall curve to provide stronger metrics, such as Mean Average Precision and Mean Average Recall with respect to the detection threshold.

% Additionally, we will develop methods of dynamically adjusting detection thresholds according to the clarity of the specific data frames. Clearer frames will have higher confidence scores than blurry frames. Therefore, adjusting accordingly will allow for the omission of poor predictions in clear videos and the admission of uncertain predictions in blurry videos, improving the reliability of detections.

%%%%%%%%% REFERENCES
{\small
\bibliographystyle{ieeetr}
\bibliography{egbib}
}

\end{document}